\documentclass{vgtc}

\usepackage{mathptmx}
\usepackage{graphicx}
\usepackage{times}
\usepackage{algorithm}
\usepackage{algpseudocode}
\usepackage{amsmath}
\usepackage{amssymb}
\usepackage{booktabs}
\usepackage{multirow}
\usepackage{threeparttable}
\usepackage{hyperref}
\usepackage{tabularx}

\onlineid{0}
\vgtccategory{Research}

\makeatletter
\renewcommand{\ps@plain}{%
    \renewcommand{\@oddhead}{}%
    \renewcommand{\@oddfoot}{\hfil\textrm{\thepage}\hfil}%
    \renewcommand{\@evenhead}{}%
    \renewcommand{\@evenfoot}{\hfil\textrm{\thepage}\hfil}%
}
\makeatother

\title{A Self‑Attentive Meta‑Optimizer\\ with Group‑Adaptive Learning Rates and Weight Decay}

\author{JiangBo Zhao
\and ZhaoXin Liu
}

\abstract{
  Adaptive optimizers like AdamW apply uniform hyperparameters across all parameter groups, ignoring heterogeneous optimization dynamics across layers and modules. We address this limitation by proposing \textbf{MetaAdamW} – a new optimizer that integrates a self‑attention mechanism to dynamically modulate per‑group learning rates and weight decay. The modulation factors are produced by a lightweight Transformer encoder that operates on statistical features (gradient norms, momentum norms, correlations) extracted from each parameter group. To train the attention module, we introduce a meta‑learning objective that combines gradient alignment, loss decrease, and generalization gap. A key novel contribution is the extension of homoscedastic uncertainty weighting \cite{kendall2018multitasklearningusinguncertainty} (HUW) with task‑specific priorities that directly scale the regularization terms – enabling domain knowledge to guide automatic loss balancing. Extensive experiments on five diverse tasks—time series forecasting (ETT), language modeling (WikiText-2), machine translation (Multi30k), image classification (CIFAR-10), and sentiment analysis (IMDB) — demonstrate that MetaAdamW consistently outperforms the standard AdamW baseline in terms of validation loss, accuracy, or perplexity. Depending on the task, MetaAdamW either reduces overall training time (by up to 17.11\%) or improves performance (by up to 11.08\%) while introducing only moderate overhead; in some cases, it can also mitigate issues of insufficient convergence caused by premature early stopping. Ablation studies validate the effectiveness of each component, including feature versions, grouping strategies, and the proposed priority‑injected uncertainty weighting.
  \smallskip
  
  \noindent \textbf{Keywords:} AdamW, Self-Attention, Meta-Learning, Group-Aware Optimization, Homoscedastic Uncertainty Weighting.
}

\nocopyrightspace

\begin{document}

\firstsection{Introduction}

\maketitle

Effective optimization algorithms are critical for deep learning success. Adam \cite{kingma2017adammethodstochasticoptimization} and its variant AdamW \cite{loshchilov2019decoupledweightdecayregularization} have become the de facto standard because of their adaptive learning rates and decoupled weight decay. However, a fundamental limitation remains: they apply identical hyperparameter settings (learning rate, weight decay) to all parameters irrespective of layer type or module function. In practice, embeddings, attention heads, and feed‑forward networks exhibit markedly different optimization dynamics. This uniformity can lead to suboptimal convergence and impaired generalization – a problem we directly target in this work.

To overcome this limitation, we introduce MetaAdamW – a principled extension of AdamW that incorporates a self‑attention mechanism for dynamic, per‑group modulation of learning rates and weight decay. Our key idea is to treat each parameter group (semantically coherent sets of parameters) as a token, extract gradient and momentum statistics as features, and apply a lightweight Transformer encoder to compute group‑specific scaling factors $\alpha$ (for learning rate) and $\beta$ (for weight decay). Unlike prior works that rely on hand‑crafted per‑layer heuristics, our method learns to modulate online from the optimization state, enabling genuinely adaptive per‑group updates.

We further equip MetaAdamW with a meta‑learning framework that periodically updates the attention module using a novel composite objective. This objective comprises three complementary terms: (i) gradient alignment between the original and a hypothetical one‑step updated model, (ii) validation loss decrease, and (iii) generalization gap (training‑validation loss difference). To automatically balance these terms, we adopt homoscedastic uncertainty weighting \cite{kendall2018multitasklearningusinguncertainty} (HUW). Crucially, we extend HUW by injecting task‑specific priorities that directly scale the regularization terms ($\log \sigma$). This priority‑injected HUW – a novel contribution – allows users to encode domain‑specific importance while retaining automatic balancing. We validate its effectiveness through ablation studies.

We evaluate MetaAdamW on five representative tasks spanning sequence modeling, vision, and language:
\begin{itemize}
    \item Time series forecasting (ETTh1) using a Transformer encoder,
    \item Language modeling (WikiText-2) using a MiniGPT model,
    \item Machine translation (Multi30k De→En) using a Transformer seq2seq,
    \item Image classification (CIFAR-10) using ResNet-18,
    \item Sentiment analysis (IMDB) using a bidirectional LSTM.
\end{itemize}
Compared to standard AdamW, MetaAdamW achieves:
\begin{itemize}
    \item For Transformer-based tasks with fewer training epochs (ETTh1, WikiText-2): lower validation loss / perplexity (improvements of 4.26\% and 4.12\%) while reducing total training time by 7.20\% and 17.11\%, respectively.
    \item For the Transformer-based Multi30k translation task (more epochs): 2.99\% lower perplexity with 27.35\% longer training time, successfully mitigating premature early stopping.
    \item For non‑Transformer architectures (ResNet‑18, LSTM): accuracy gains of 1.18\% and 11.08\%, respectively, with increased training time (27.58\% and 172.53\%), again avoiding too‑early stopping.
\end{itemize}
These results demonstrate that MetaAdamW not only improves generalization but also offers flexible trade‑offs between performance and training cost depending on task characteristics.

Ablation studies further dissect the contributions of feature versions, grouping strategies, meta-objectives, and the proposed priority-injected uncertainty weighting. Our results confirm that the attention-based modulation and meta-learning are effective and generalizable across different architectures and tasks.

\section{Related Work}

\subsection{Adaptive Optimization Algorithms}
The Adam optimizer \cite{kingma2017adammethodstochasticoptimization} combines momentum and RMSProp, providing adaptive learning rates per parameter. AdamW \cite{loshchilov2019decoupledweightdecayregularization} decouples weight decay from the adaptive updates, improving generalization. Despite their success, these methods treat all parameters uniformly. Several works have attempted to introduce per-layer or per-parameter hyperparameters. For instance, HyperAdam \cite{wang2018hyperadamlearnabletaskadaptiveadam} employs a recurrent meta-network to generate adaptive decay rates and combination weights, requiring an additional meta-optimization loop. Meta-SGD \cite{li2017metasgdlearninglearnquickly} learns a parameter-wise learning rate vector and initialization via nested gradient-based meta-learning, but does not capture interactions among parameter groups. In contrast, MetaAdamW leverages self-attention to capture dependencies among parameter groups and computes modulation factors in a single forward pass, without introducing a separate meta-network.

\subsection{Learning to Optimize with Recurrent Networks}
Transformers and self-attention \cite{vaswani2023attentionneed} have revolutionized sequence modeling. Recent works have explored learning to optimize using recurrent neural networks (e.g., LSTMs) \cite{andrychowicz2016learninglearngradientdescent} or hypernetwork-based adaptive optimizers \cite{wang2018hyperadamlearnabletaskadaptiveadam}. However, these approaches either rely on hand-designed update rules (HyperAdam) or require a separate meta-optimizer with high computational cost. In contrast, MetaAdamW integrates a lightweight Transformer encoder directly into the optimizer to compute per-group modulation factors, leveraging self-attention to capture dependencies among parameter groups while incurring minimal overhead.

\subsection{Meta-Learning for Optimization}
Meta-learning aims to learn learning algorithms themselves. In the context of optimization, MAML \cite{finn2017modelagnosticmetalearningfastadaptation} learns a model initialization that enables fast adaptation with few gradient steps. Another line of work, such as the LSTM-based optimizer \cite{andrychowicz2016learninglearngradientdescent} (often termed "learning to learn by gradient descent") and Meta-SGD \cite{li2017metasgdlearninglearnquickly}, trains a meta-learner to directly produce parameter updates. Our meta-update scheme is inspired by these approaches but is tailored to the attention module within the optimizer. We adopt a bilevel optimization formulation where the inner loop is a single MetaAdamW step, and the outer loop minimizes a combined objective computed on separate mini-batches.

\subsection{HUW and Priority Injection}
Multi-task learning often requires balancing multiple loss terms. Kendall et al. \cite{kendall2018multitasklearningusinguncertainty} proposed using homoscedastic uncertainty (learnable task variances) to automatically weight losses. For a regression task, the log-likelihood is $-\frac{1}{2\sigma^2}|\mathbf{y}-\hat{\mathbf{y}}|^2 - \log\sigma$; for classification, a scaled softmax is used. The learnable $\sigma$ (or log-variance) automatically balances the losses. In this work, we extend this formulation by introducing \textit{task-specific priorities} $p_i$ that scale the regularization term (but not the data term). The resulting loss for a regression task becomes $\sum_i \left( \frac{1}{2\sigma_i^2}\mathcal{L}_i + p_i \log\sigma_i \right)$, and for a classification task becomes $\sum_i \left( \frac{1}{\sigma_i^2}\mathcal{L}_i + p_i \log\sigma_i^2 \right)$, following the approximation in \cite{kendall2018multitasklearningusinguncertainty} and consistent with our implementation.

\section{Methodology}

\subsection{Problem Formulation}
Consider a deep neural network with parameters $\theta$. Standard AdamW maintains first and second moments for each parameter. Let the parameter set be partitioned into $G$ groups $\{\mathcal{P}_g\}_{g=1}^G$ according to a grouping strategy (Sec.~\ref{subsec:grouping}). For each group $g$, we aim to compute two modulation factors: $\alpha_g$ (scaling the learning rate) and $\beta_g$ (scaling the weight decay). The update rule for a parameter $p \in \mathcal{P}_g$ becomes:
\begin{align}
    m_t &= \beta_1 m_{t-1} + (1-\beta_1) \nabla L(\theta_{t-1}), \\
    v_t &= \beta_2 v_{t-1} + (1-\beta_2) (\nabla L)^2, \\
    \hat{m}_t &= m_t / (1-\beta_1^t), \quad \hat{v}_t = v_t / (1-\beta_2^t), \\
    \theta_t &= \theta_{t-1} - \eta \alpha_g \left( \frac{\hat{m}_t}{\sqrt{\hat{v}_t}+\epsilon} + \beta_g \lambda \theta_{t-1} \right),
\end{align}
where $\eta$ is the global learning rate, $\lambda$ the weight decay, and $\epsilon$ a small constant.

The core challenge is to compute $\alpha_g$ and $\beta_g$ dynamically based on the current optimization state.

\subsection{Group Construction}
\label{subsec:grouping}
We implement a fine‑grained grouping strategy that categorizes parameters by layer type (embedding, attention, feed‑forward, layer norm, other), depth bucket (shallow, middle, deep), and bias indicator. For the attention module, this grouping method represents a manageable scale. For models where internal structure is unavailable, we fall back to PyTorch’s native parameter groups.

\subsection{Feature Extraction}
For each group $\mathcal{P}_g$, we extract a feature vector $\mathbf{f}_g \in \mathbb{R}^{D}$ that summarizes the optimization dynamics. We propose three core feature versions:
\begin{itemize}
    \item \textbf{Basic}: 4 statistics (mean of gradient norms, momentum norms, parameter norms, cosine similarity between gradient and momentum) plus optional time step and optional second-order momentum norm.
    \item \textbf{Basic+}: adds standard deviations of the four basic statistics.
    \item \textbf{Enhanced}: 9 statistics (mean and variance of gradient norms, mean and variance of momentum norms, gradient sparsity, momentum sparsity, log parameter count, bias ratio, normalized layer depth) plus time step and a learnable group embedding.
\end{itemize}
Additionally, we provide normalized variants that perform cross‑group standardization on the statistical columns.

\subsection{Attention-Based Modulation}
Let $\mathbf{F} = [\mathbf{f}_1; \dots; \mathbf{f}_G] \in \mathbb{R}^{G \times D}$ be the feature matrix. We apply a Transformer encoder with $L$ layers, $H$ heads, and hidden dimension $D_{\text{ff}}$:
\begin{equation}
    \mathbf{Z} = \text{TransformerEncoder}(\mathbf{F}), \quad \mathbf{Z} \in \mathbb{R}^{G \times D}.
\end{equation}
A linear projection outputs four raw values per group:
\begin{equation}
    [\alpha_g^{\text{raw}}, \beta_g^{\text{raw}}, \lambda_{1,g}^{\text{raw}}, \lambda_{2,g}^{\text{raw}}]^\top = \mathbf{W}_{\text{out}} \mathbf{z}_g + \mathbf{b}_{\text{out}}.
\end{equation}
The final modulation factors are:
\begin{align}
    \alpha_g &= 1 + \text{range}_\alpha \cdot (\sigma(\alpha_g^{\text{raw}}) - 0.5), \\
    \beta_g  &= 1 + \text{range}_\beta  \cdot (\sigma(\beta_g^{\text{raw}}) - 0.5), \\
    \lambda_{1,g} &= \sigma(\lambda_{1,g}^{\text{raw}}), \quad \lambda_{2,g} = \sigma(\lambda_{2,g}^{\text{raw}}),
\end{align}
where $\sigma$ is the sigmoid function. $\lambda_{1,g}$ and $\lambda_{2,g}$ are used in the meta-learning objective (Sec.~\ref{subsec:meta}).

\subsection{Meta-Learning Objectives}
\label{subsec:meta}
We periodically update the attention module (every $K_{\text{meta}}$ steps) to minimize a combined objective that promotes desirable optimization properties. The meta-update uses two separate mini-batches $\mathcal{B}_1$ and $\mathcal{B}_2$ (drawn from the training set) and a validation batch $\mathcal{B}_{\text{val}}$.

Let $\theta$ be the model parameters, and let $\theta'$ denote the parameters after one MetaAdamW step using the current attention module. The meta-loss consists of three terms:
\begin{enumerate}
    \item \textbf{Gradient alignment} ($\mathcal{L}_{\text{grad}}$): Encourages the gradient of the updated model on $\mathcal{B}_2$ to align with the original gradient on $\mathcal{B}_1$:
    \begin{equation}
        \mathcal{L}_{\text{grad}} = \frac{1}{G}\sum_{g=1}^G \left( \lambda_{1,g} \|\mathbf{g}_g'\|^2 - \lambda_{2,g} \cos(\mathbf{g}_g, \mathbf{g}_g') \right),
    \end{equation}
    where $\mathbf{g}_g$ is the aggregated gradient of group $g$ on the first mini‑batch $\mathcal{B}_1$, and $\mathbf{g}_g'$ is the aggregated gradient of the same group on $\mathcal{B}_2$ \textit{after} the hypothetical one‑step update.
    \item \textbf{Loss decrease} ($\mathcal{L}_{\text{loss}}$): Measures the reduction in validation loss:
    \begin{equation}
        \mathcal{L}_{\text{loss}} = L_{\text{val}}(\theta') - L_{\text{val}}(\theta).
    \end{equation}
    \item \textbf{Generalization gap} ($\mathcal{L}_{\text{gap}}$): Penalizes the difference between training and validation losses:
    \begin{equation}
        \mathcal{L}_{\text{gap}} = |L_{\text{train}}(\theta, \mathcal{B}_1) - L_{\text{val}}(\theta)|.
    \end{equation}
\end{enumerate}

Meta-learning objectives can be selected from four options: Gradient alignment, Loss decrease, Generalization gap, and Combined.

When the meta-learning objective is Combined, instead of fixed weights, we learn the weights adaptively using homoscedastic uncertainty \cite{kendall2018multitasklearningusinguncertainty}. For each term $i$, we introduce a learnable log-variance $s_i = \log \sigma_i^2$. Since all three meta-loss terms ($\mathcal{L}{\text{grad}}$, $\mathcal{L}{\text{loss}}$, $\mathcal{L}{\text{gap}}$) are scalar regression objectives, we adopt the regression formulation. The meta‑loss is defined as:
\begin{equation}
\mathcal{L}{\text{meta}} = \sum_{i=1}^3 \left( \frac{1}{2\sigma_i^2} \mathcal{L}_i + p_i \cdot \frac{1}{2} \log \sigma_i^2 \right),
\end{equation}
where $p_i$ is a user‑defined priority for task $i$. (For completeness, a classification task would use $\frac{1}{\sigma_i^2} \mathcal{L}_i + p_i \log \sigma_i^2$ instead, following our implementation.) To the best of our knowledge, this priority injection into the HUW has not been previously proposed; it offers a flexible way to incorporate domain knowledge while preserving automatic balancing.

\subsection{Algorithm Summary}
The optimizer maintains standard AdamW states and additionally updates the attention module every $K_{\text{meta}}$ steps via the meta-objective. In practice, the meta‑update is not embedded inside the main \texttt{step()} function; rather, the user is responsible for invoking \texttt{update\_attention()} at the desired frequency after the warmup phase. Algorithm~1 outlines the logical flow of a single training iteration, assuming the meta‑update is triggered externally when the condition holds.

Algorithm~\ref{alg:meta_adamw} outlines the overall procedure. The optimizer maintains standard AdamW states and additionally updates the attention module every $K_{\text{meta}}$ steps via the meta-objective.

\begin{algorithm}[t]
\caption{MetaAdamW Update (one step)}
\label{alg:meta_adamw}
\begin{algorithmic}[1]
\Require Model $\theta$, learning rate $\eta$, weight decay $\lambda$, meta-update frequency $K_{\text{meta}}$, global step $t$.
\Ensure Updated $\theta$, attention module $\Phi$.
\State \Comment{Standard AdamW states: $m$, $v$}
\State Compute groups $\{\mathcal{P}_g\}$ and extract features $\{\mathbf{f}_g\}$.
\State Compute modulation factors $\alpha_g, \beta_g$ via $\Phi$ (attention encoder + output projection).
\For{each parameter $p$ in group $g$}
    \State Update $m_t, v_t$ using current gradient.
    \State Apply update: $p \gets p - \eta \alpha_g ( \frac{\hat{m}_t}{\sqrt{\hat{v}_t}+\epsilon} + \beta_g \lambda p)$.
\EndFor
\If{$t \bmod K_{\text{meta}} = 0$ and $t \geq \text{warmup\_steps}$}
    \State \textbf{Note:} In practice, the meta‑update is not executed automatically inside \texttt{step()}; instead, the training loop must explicitly call \texttt{update\_attention()} every $K_{\text{meta}}$ steps after the warmup period. The pseudocode below assumes this external invocation.
    \State Perform meta-update of $\Phi$ using two training batches and a validation batch (Algorithm~\ref{alg:meta_update}).
\EndIf
\State $t \gets t+1$
\end{algorithmic}
\end{algorithm}

The key steps of the meta-update algorithm (Appendix~\ref{app:algo_meta_update}) are:

\begin{itemize}
    \item \textbf{Save original parameters}: First, the original model parameters are saved;
    \item \textbf{Compute original gradients}: then, gradients are computed on the first mini‑batch $\mathcal{B}_1$ using the current model.
    \item \textbf{Compute modulation factors}: Modulation factors $(\alpha_{\text{g}}, \beta_{\text{g}}, \lambda_{\text{1,g}}, \lambda_{\text{2,g}})$ are obtained from the attention module.
    \item \textbf{Construct hypothetical parameters}: Hypothetical parameters $\theta'$ are constructed by applying one MetaAdamW step.
    \item \textbf{Compute gradients on hypothetical parameters}: Gradients $\mathbf{g}'$ are computed on $\theta'$ using the second mini‑batch $\mathcal{B}_2$.
    \item \textbf{Compute meta‑loss}: A meta‑loss $\mathcal{L_{\text{meta}}}$ (combining gradient alignment, loss decrease, and generalization gap, weighted by priority‑injected homoscedastic uncertainty) is evaluated,
    \item \textbf{Backpropagate to update attention module}: and backpropagated through the attention module to update its parameters.
    \item \textbf{Restore original parameters}: Finally, the original model parameters are restored.
\end{itemize}

\section{Experiments}

\subsection{Experimental Setup}

\subsubsection{Tasks and Datasets}
To validate the effectiveness of MetaAdamW and to demonstrate its generality, we conduct experiments on five fundamentally different tasks:
\begin{itemize}
    \item \textbf{Time Series Forecasting (ETT)}: ETTh1 dataset (hourly, 7 features). Input length 96, predict next 1 step (oil temperature). Model: Transformer encoder with $d_{\text{model}}=128$, 2 layers, 4 heads.
    \item \textbf{Language Modeling (WikiText-2)}: Word-level language modeling with block size 128, vocabulary size $\approx$10k. Model: MiniGPT with $d_{\text{model}}=128$, 2 layers, 4 heads.
    \item \textbf{Machine Translation (Multi30k)}: German→English translation, sequence length 64. Model: Transformer seq2seq with $d_{\text{model}}=256$, 4 encoder/decoder layers, 8 heads.
    \item \textbf{Image Classification (CIFAR-10)}: 10-class classification. Model: ResNet-18 (modified for 32$\times$32 inputs).
    \item \textbf{Sentiment Analysis (IMDB)}: Binary sentiment classification, sequence length 256. Model: Bidirectional LSTM with embedding dimension 32, hidden size 64, 2 layers.
\end{itemize}

\subsubsection{Baseline and Hyperparameters}
We compare MetaAdamW against standard AdamW (decoupled weight decay). For fairness, both optimizers use the same base learning rate $5e-4$, weight decay $1e-2$, and betas $(0.9,0.999)$. MetaAdamW hyperparameters that are common across tasks are listed in Table~\ref{tab:hyper}. Task‑specific optimal hyperparameters (e.g., meta‑update frequency, attention depth, feature version, HUW priorities) are provided in Appendix~\ref{app:task_hyper}.

\begin{table}[htb]
\centering
\caption{MetaAdamW Common Hyperparameters}
\label{tab:hyper}
\begin{tabular*}{\columnwidth}{@{\extracolsep{\fill}} l r}
\toprule
Parameter & Value \\
\midrule
Global learning rate $\eta$ & $5\times10^{-4}$ \\
Weight decay $\lambda$ & $1\times10^{-2}$ \\
AdamW betas & $(0.9, 0.999)$ \\
$\alpha$ range & $1.0$ \\
$\beta$ range & $1.0$ \\
Warmup epochs (before meta) & $1$ \\
Group strategy & fine-grained \\
Feature version & basic (or basic\_plus for LSTM) \\
Use $v$ norms & True \\
Include time step & True \\
Meta objective & combined \\
\bottomrule
\end{tabular*}
\end{table}

\subsubsection{Evaluation Metrics}
We report:
\begin{itemize}
    \item \textbf{Validation loss} - ETT: MSE; LM: cross-entropy; MT: cross-entropy; CIFAR/IMDB: cross-entropy.
    \item \textbf{Task-specific metric} - ETT: MSE; LM: perplexity; MT: perplexity; CIFAR/IMDB: accuracy.
    \item \textbf{Total training time} - (seconds) and relative change.
\end{itemize}

Early stopping with patience 2 is applied for both optimizers.

\subsection{Results}

\subsubsection{Performance Comparison}
Table~\ref{tab:main_results} reports the best validation metrics and total training time for AdamW and MetaAdamW on all five tasks. MetaAdamW consistently outperforms the baseline on every task, with improvements ranging from modest (1.18\% accuracy on CIFAR-10) to substantial (11.08\% accuracy on IMDB). Notably, for two Transformer-based tasks (ETTh1, WikiText-2), MetaAdamW achieves better performance (4.26\% and 4.12\%) while also reducing total training time (by 7.20\% and 17.11\%, respectively), because it reaches a better optimum earlier and triggers early stopping sooner. For the more complex Multi30k translation task, MetaAdamW avoids premature early stopping (training for more epochs) and yields a 2.99\% perplexity reduction at a 27.35\% time cost. For non‑Transformer architectures (ResNet-18, LSTM), MetaAdamW also trains longer (27.58\% and 172.53\%) but delivers clear accuracy gains (1.18\% and 11.08\%).

\begin{table*}[htb]
\centering
\caption{Main Results: Best Validation Metrics and Total Training Time}
\label{tab:main_results}
\begin{tabular*}{\textwidth}{@{\extracolsep{\fill}} l l r l r r r}
\toprule
\multirow{2}{*}{Task} & \multicolumn{2}{c}{AdamW} & \multicolumn{2}{c}{MetaAdamW} & \multirow{2}{*}{Improvement} & \multirow{2}{*}{Time Overhead} \\
\cmidrule(lr){2-3} \cmidrule(lr){4-5}
 & Metric & Value & Metric & Value & & \\
\midrule
ETTh1 (forecast) & MSE & 0.006147 & MSE & 0.005885 & +4.26\% & -7.20\% \\
WikiText-2 (LM) & PPL & 120.47 & PPL & 115.51 & +4.12\% & -17.11\% \\
Multi30k (MT) & PPL & 2.0297 & PPL & 1.9690 & +2.99\% & +27.35\% \\
CIFAR-10 (classif.) & Acc & 87.02\% & Acc & 88.05\% & +1.18\% & +27.58\% \\
IMDB (sentiment) & Acc & 74.53\% & Acc & 82.79\% & +11.08\% & +172.53\% \\
\bottomrule
\end{tabular*}
\end{table*}

\begin{figure}[htb]
  \centering
  \includegraphics[width=1.0\columnwidth]{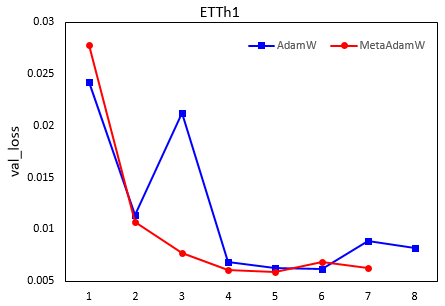}
  \caption{ETTh1 Validation loss over epochs.}
  \label{fig:ETTh1}
\end{figure}

\begin{figure}[htb]
  \centering
  \includegraphics[width=1.0\columnwidth]{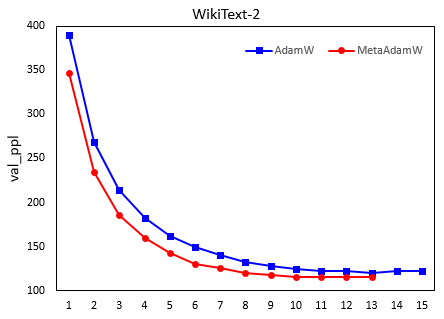}
  \caption{WikiText-2 Validation perplexity over epochs.}
  \label{fig:WikiText-2}
\end{figure}

\begin{figure}[htb]
  \centering
  \includegraphics[width=1.0\columnwidth]{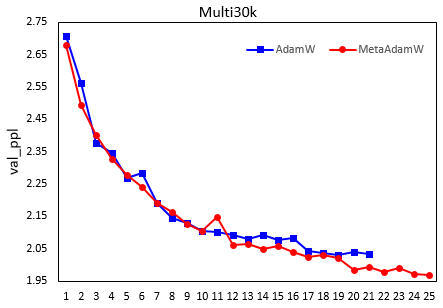}
  \caption{Multi30k Validation perplexity over epochs.}
  \label{fig:Multi30k}
\end{figure}

\begin{figure}[htb]
  \centering
  \includegraphics[width=1.0\columnwidth]{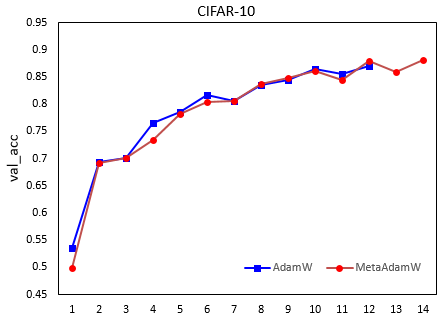}
  \caption{CIFAR-10 Validation accuracy over epochs.}
  \label{fig:CIFAR-10}
\end{figure}

\begin{figure}[htb]
  \centering
  \includegraphics[width=1.0\columnwidth]{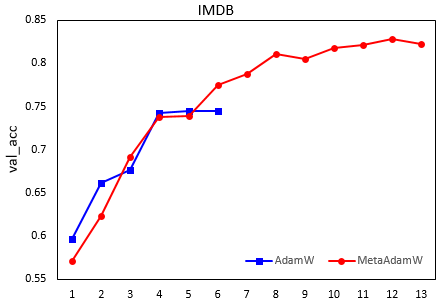}
  \caption{IMDB Validation accuracy over epochs.}
  \label{fig:IMDB}
\end{figure}

Figure~\ref{fig:ETTh1}, ~\ref{fig:WikiText-2}, ~\ref{fig:Multi30k}, ~\ref{fig:CIFAR-10} and ~\ref{fig:IMDB} show the validation curves for each task. MetaAdamW may converge faster, or ultimately perform better, or help avoid premature stopping and achieve better convergence, depending on the task.

\subsubsection{Computational Overhead}
The additional cost of MetaAdamW comes from feature extraction, attention forward/backward, and meta-updates. As shown in Table~\ref{tab:main_results}, the total training time change varies: it decreases for ETTh1 and WikiText-2 (due to earlier stopping) but increases for the other three tasks. The largest relative increase occurs on IMDB (172.5\%), because the LSTM model has many parameter groups and the meta-update frequency was set high (every 39 steps). Nevertheless, the performance gains justify the extra cost in those scenarios.

\subsection{Ablation Studies}

To systematically assess the contribution of each design component, we conducted extensive ablations across all five tasks. Table~\ref{tab:ablation_summary} reports representative results on five tasks. All hyperparameters except the ablated one follow the task‑optimal settings (Appendix~\ref{app:task_hyper}).

\begin{table*}[htb]
\centering
\caption{Ablation study: validation performance changes.}
\label{tab:ablation_summary}
\begin{tabular*}{\textwidth}{@{\extracolsep{\fill}} l r r r r r r r r r r}

\toprule
\multirow{2}{*}{Configuration} & \multicolumn{2}{c}{ETTh1 (Loss)} & \multicolumn{2}{c}{WikiText-2 (PPL)} & \multicolumn{2}{c}{Multi30k (PPL)} & \multicolumn{2}{c}{CIFAR-10 (Acc)} & \multicolumn{2}{c}{IMDB (Acc)} \\

\cmidrule(lr){2-3} \cmidrule(lr){4-5} \cmidrule(lr){6-7} \cmidrule(lr){8-9} \cmidrule(lr){10-11}
& Value & ${\Delta}$\% & Value & ${\Delta}$\% & Value & ${\Delta}$\% & Value & ${\Delta}$\% & Value & ${\Delta}$\% \\

\midrule
MetaAdamW (baseline) & 0.005885 & – & 115.51 & – & 1.9690 & – & 0.8805 & – & 0.8279 & – \\

\hline
AdamW & 0.006147 & -4.45 & 120.47 & -4.29 & 2.0297 & -3.08 & 0.8702 & -1.17 & 0.7453 & -9.98 \\

\hline
\textbf{Grouping strategy} & & & & & & & & & & \\
~~Original & – & – & 116.13 & -0.54 & – & – & – & – & – & – \\
~~Fine-grained & baseline & – & baseline & – & baseline & – & baseline & – & baseline & – \\

\hline
\textbf{Feature version} & & & & & & & & & & \\
~~Basic & baseline & – & baseline & – & baseline & – & baseline & – & 0.7938 & -4.12 \\
~~Basic plus & – & – & 117.78 & -1.97 & 2.6180 & -32.96 & – & – & baseline & – \\
~~Enhanced & – & – & 121.02 & -4.77 & – & – & – & – & – & – \\

\hline
\textbf{Meta-learning objective} & & & & & & & & & & \\
~~Gradient & – & – & 116.13 & -0.54 & – & – & 0.8304 & -5.69 & – & – \\
~~Loss decrease & – & – & 116.70 & -1.03 & – & – & – & – & – & – \\
~~Generalization gap & – & – & 116.47 & -0.83 & – & – & – & – & – & – \\
~~Combined & baseline & – & baseline & – & baseline & – & baseline & – & baseline & – \\

\hline
\textbf{HUW priorities} & & & & & & & & & & \\
~~No HUW & – & – & 116.13 & -0.54 & – & – & – & – & – & – \\
~~[1.0, 1.0, 1.0] & baseline & – & baseline & – & 2.1205 & -7.69 & 0.8602 & -2.31 & – & – \\
~~Not [1.0, 1.0, 1.0] & – & – & 115.57 & -0.05 & baseline & – & baseline & – & baseline & – \\

\hline
\textbf{Feature gating} & & & & & & & & & & \\
~~Enable & baseline & – & 116.17 & -0.57 & 2.6180 & -32.96 & 0.8602 & -2.31 & baseline & – \\
~~Disable & 0.005932 & -0.80 & baseline & – & baseline & – & baseline & – & 0.7938 & -4.12 \\

\hline
\textbf{Meta-update frequency} & & & & & & & & & & \\
~~Once per epoch & baseline & – & baseline & – & baseline & – & 0.8776 & -0.33 & 0.6767 & -18.26 \\
~~Approx. twice per epoch & – & – & 116.13 & -0.54 & – & – & baseline & – & – & – \\
~~Approx. 10 times per epoch & – & – & 116.52 & -0.87 & 2.0338 & -3.29 & – & – & baseline & – \\

\hline
\textbf{Attention Module} & & & & & & & & & & \\
~~${L_{\text{attn}}}$=8, ${D_{\text{attn}}}$=16 & baseline & – & baseline & – & 2.0338 & -3.29 & 0.8579 & -2.57 & – & – \\
~~Not (${L_{\text{attn}}}$=8, ${D_{\text{attn}}}$=16) & – & – & 116.32 & -0.70 & baseline & – & baseline & – & baseline & – \\

\bottomrule
\end{tabular*}
\textit{Notes: (1) ${\Delta}$\% indicates relative change from baseline; negative ${\Delta}$\% means worse performance. (2) “–” means no data or baseline.}
\end{table*}

\paragraph{Grouping strategy.}
Fine‑grained grouping (by layer type, depth, and bias) consistently outperforms PyTorch’s native parameter groups.

\paragraph{Feature version.}
The simple \texttt{basic} feature set (mean of gradient / momentum / parameter norms and cosine similarity, plus time step and second‑moment norm) achieves near‑optimal results. Adding standard deviations (\texttt{basic\_plus}) hurts performance. The more complex \texttt{enhanced} features perform even worse. Thus, \texttt{basic} is sufficient for most tasks.

\paragraph{Meta-learning objective.}
Removing the attention modulation (i.e., standard AdamW) causes substantial degradation. The \texttt{combined} objective outperforms any single component.

\paragraph{HUW priorities.}
Priority‑injected homoscedastic uncertainty weighting (HUW) improves over fixed equal weights. This validates the benefit of task‑specific priority scaling.

\paragraph{Feature gating.}
Learnable feature gating still requires further research and exploration.

\paragraph{Meta-update frequency.}
An excessively high meta-update frequency significantly increases computational time and does not necessarily yield substantial performance improvements; to a large extent, this depends on the specific task.

\paragraph{Attention module size.}
More complex tasks benefit from greater expressivity.

In summary, the ablation study confirms that the substantial performance of self‑attentive MetaAdamW optimizer stems from the synergy of fine‑grained grouping, the combined meta‑objective, and the priority‑injected HUW balancing.

\subsection{Discussion}

The results demonstrate that MetaAdamW effectively learns to modulate per-group learning rates and weight decay. The meta-learning component adapts the attention module to the specific task and training dynamics, leading to better generalization. The computational overhead is task‑dependent: for tasks where early stopping is triggered earlier (ETTh1, WikiText-2), total time actually decreases; for others, the extra time is compensated by improved accuracy or perplexity.

A notable observation is that the optimal hyperparameters of MetaAdamW vary substantially across tasks. For instance, the LSTM sentiment task required a higher attention head count (11 vs. 6) and the \texttt{basic\_plus} feature version, while the Transformer-based tasks worked well with the simpler \texttt{basic} features. The meta‑update frequency also differed: 217 steps for ETTh1, 123 for WikiText-2, 454 for Multi30k, 190 for CIFAR-10, and 39 for IMDB. This diversity underscores the flexibility of MetaAdamW and suggests that practitioners may need to tune a few key parameters for their specific domain. We provide the full set of optimal configurations in Appendix~\ref{app:task_hyper} to facilitate reproducibility.

\subsection{Memory and Computation Overhead}

We analyze the peak memory and computational cost of MetaAdamW compared to the baseline AdamW, as this is critical for practical deployment.

\textbf{Peak memory footprint.}  
Standard AdamW stores model parameters ($P$), gradients ($P$), first‑moment ($P$) and second‑moment ($P$) buffers, plus forward activations and the backward computation graph. Its peak memory is approximately $3P\text{--}4P + \text{activations} + \text{graph}$.

MetaAdamW introduces an additional peak during meta‑update steps (every $K_{\text{meta}}$ iterations). In such a step, the following coexist:
\begin{itemize}
    \item Original model parameters, gradients (`grads1`), and momentum buffers (total $\sim 4P$)
    \item Temporary parameters `temp\_params` (a full copy of the model, size $P$)
    \item Second‑order gradients `grads\_temp` (size $P$)
    \item Two sets of forward activations (for batches 1 and 2)
    \item A second backward computation graph
\end{itemize}
Consequently, the \textbf{peak memory during a meta‑update} can reach roughly $5P\text{--}6P + 2\times\text{activations} + \text{graph}_2$, i.e., \textbf{1.5–2$\times$} the peak of AdamW. During regular (non‑meta) steps, MetaAdamW's peak memory is nearly identical to AdamW because no temporary parameters or second backward pass are created.

\textbf{Computational cost.}  
In a regular step, feature extraction adds about 0.5-1$\times$ the FLOPs of the baseline AdamW update, while the attention module contributes negligible overhead. Meta‑updates require an extra full forward‑backward pass ($\sim 1\times$ model computation) every $K_{\text{meta}}$ steps. With $K_{\text{meta}}=100$, the average extra computation is roughly 1\% of model computation plus the per‑step feature extraction cost. In practice, end‑to‑end training time \textbf{increases by 10–30\%}.

\subsection{Future Work}

While MetaAdamW is validated on lightweight models, scaling it to billion‑parameter Transformers and more complex tasks remains an important direction for further investigation. The current implementation incorporates a feature gating mechanism, which applies a learnable gate to each feature — mediated by a Sigmoid activation function — and is subsequently complemented by L1 sparsity regularization. Its effectiveness has been demonstrated in the current experiments; we will conduct further in-depth research into the feature gating mechanism.

\section{Conclusion}

We have introduced MetaAdamW, a new optimizer that integrates self‑attention and meta‑learning into the AdamW framework. It solves the problem of uniform hyperparameter treatment by extracting statistical features from parameter groups and using a Transformer encoder to compute adaptive, per‑group learning rates and weight decay. A novel meta‑learning objective – combining gradient alignment, loss decrease, and generalization gap – balances these terms via priority‑injected homoscedastic uncertainty weighting, a mechanism we propose and validate. Our experiments across five diverse tasks (ETTh1, WikiText-2, Multi30k, CIFAR-10, IMDB) show that MetaAdamW consistently outperforms standard AdamW. Depending on the task, it either reduces total training time (by up to 17.11\%) or achieves better performance with modest overhead (e.g., 2.99\% lower perplexity on Multi30k, 11.08\% higher accuracy on IMDB) while mitigating premature early stopping. Ablation studies confirm the necessity of each design component.

\bibliographystyle{abbrv}
\bibliography{references}

\appendix

\section{Meta-Update Algorithm}
\label{app:algo_meta_update}
Algorithm~\ref{alg:meta_update} computes the hypothetical parameters $\theta'$ using current $\Phi$, evaluates $\mathcal{L}_{\text{meta}}$, and backpropagates through the attention module.

\begin{algorithm*}[t]
\caption{Meta-Update of Attention Module}
\label{alg:meta_update}
\begin{algorithmic}[1]
\Require Model $\theta$ (current parameters), attention module $\Phi$ (including group embeddings), two training batches $\mathcal{B}_1, \mathcal{B}_2$, validation batch $\mathcal{B}_{\text{val}}$ (if needed), loss function $\mathcal{L}_{\text{task}}$, meta-learning rate $\eta_{\text{meta}}$, current optimizer states $\{m,v\}$.
\Ensure Updated attention module $\Phi$.

\State \Comment{Step 1: Save original parameters}
\For{each parameter $p$ in $\theta$}
    \State $p_{\text{orig}} \gets p.\text{data.clone}()$
\EndFor

\State \Comment{Step 2: Compute original gradients $\mathbf{g}$ on $\mathcal{B}_1$}
\State $\mathcal{L}_1 \gets \mathcal{L}_{\text{task}}(\theta, \mathcal{B}_1)$
\State $\mathbf{g} \gets \nabla_{\theta} \mathcal{L}_1$ \Comment{gradients w.r.t. all parameters}

\State \Comment{Step 3: Compute modulation factors using current $\Phi$}
\State Extract group features $\{\mathbf{f}_g\}$ from current states and time step.
\State $(\alpha_g, \beta_g, \lambda_{1,g}, \lambda_{2,g}) \gets \Phi(\{\mathbf{f}_g\})$ \Comment{as in Algorithm~1}

\State \Comment{Step 4: Build hypothetical parameters $\theta'$ after one AdamW step}
\For{each parameter group $g$}
    \For{each parameter $p \in \mathcal{P}_g$}
        \State Compute $m', v'$ using standard AdamW update (Eq.~1--3) with $\alpha_g, \beta_g$.
        \State $\theta'_p \gets p - \eta \alpha_g \big( \frac{\hat{m}'}{\sqrt{\hat{v}'}+\epsilon} + \beta_g \lambda p \big)$
    \EndFor
\EndFor

\State \Comment{Step 5: Compute gradients $\mathbf{g}'$ on $\mathcal{B}_2$ using $\theta'$}
\State $\mathcal{L}_2 \gets \mathcal{L}_{\text{task}}(\theta', \mathcal{B}_2)$
\State $\mathbf{g}' \gets \nabla_{\theta'} \mathcal{L}_2$

\State \Comment{Step 6: Compute meta-loss $\mathcal{L}_{\text{meta}}$}
\State $\mathcal{L}_{\text{grad}} \gets \frac{1}{G}\sum_{g=1}^G \big( \lambda_{1,g}\|\mathbf{g}_g'\|^2 - \lambda_{2,g} \cos(\mathbf{g}_g, \mathbf{g}_g') \big)$
\If{meta objective includes loss decrease}
    \State $\mathcal{L}_{\text{loss}} \gets \mathcal{L}_{\text{task}}(\theta', \mathcal{B}_{\text{val}}) - \mathcal{L}_{\text{task}}(\theta, \mathcal{B}_{\text{val}})$
\EndIf
\If{meta objective includes generalization gap}
    \State $\mathcal{L}_{\text{gap}} \gets |\mathcal{L}_{\text{task}}(\theta, \mathcal{B}_1) - \mathcal{L}_{\text{task}}(\theta, \mathcal{B}_{\text{val}})|$
\EndIf
\State $\mathcal{L}_{\text{meta}} \gets \text{CombinedObjective}(\mathcal{L}_{\text{grad}}, \mathcal{L}_{\text{loss}}, \mathcal{L}_{\text{gap}})$ \Comment{weighted by HUW with priorities}

\State \Comment{Step 7: Backpropagate through $\Phi$ and update}
\State $\nabla_{\Phi} \mathcal{L}_{\text{meta}} \gets \text{autograd}(\mathcal{L}_{\text{meta}})$
\For{each parameter $\phi$ in $\Phi$ (including group embeddings)}
    \State $\phi \gets \phi - \eta_{\text{meta}} \nabla_{\phi} \mathcal{L}_{\text{meta}}$
\EndFor

\State \Comment{Step 8: Restore original parameters}
\For{each parameter $p$ in $\theta$}
    \State $p.\text{data} \gets p_{\text{orig}}$
\EndFor
\end{algorithmic}
\end{algorithm*}

\section{Task-Specific Hyperparameters}
\label{app:task_hyper}
Table~\ref{tab:task_hyper} reports the optimal hyperparameter configuration for each task, as discovered during our ablation sweeps. These values can serve as a starting point for future applications.

\begin{table*}[htb]
\centering
\caption{Optimal MetaAdamW hyperparameters per task}
\label{tab:task_hyper}
\begin{tabular*}{\textwidth}{@{\extracolsep{\fill}} l r r r r r r r r}
\toprule
Task & $K_{\text{meta}}$ & $L$ (attn layers) & $D_{\text{hid}}$ (attn) & Heads & Feature version & Feature dim & Feature gating & HUW priorities \\
\midrule
ETTh1 & 217 & 8 & 16 & 6 & basic & 6 & True & [1.0, 1.0, 1.0] \\
WikiText-2 & 123 & 8 & 16 & 6 & basic & 6 & False & [1.0, 1.0, 1.0] \\
Multi30k & 454 & 128 & 64 & 6 & basic & 6 & False & [2.0, 5.0, 1.0] \\
CIFAR-10 & 190 & 64 & 64 & 6 & basic & 6 & False & [5.0, 2.0, 1.0] \\
IMDB & 39 & 64 & 64 & 11 & basic\_plus & 11 & True & [1.0, 3.0, 5.0] \\
\bottomrule
\end{tabular*}
\textit{Note: The HUW priorities correspond to the three meta-loss terms in the order: gradient alignment, loss decrease, generalization gap.}
\end{table*}

\section{Detailed Training Time Breakdown}
Table~\ref{tab:time} shows the average epoch time (seconds) and total training time for AdamW vs. MetaAdamW.

\begin{table*}[htb]
\centering
\caption{Average epoch time (seconds) and total training time}
\label{tab:time}
\begin{tabular*}{\textwidth}{@{\extracolsep{\fill}} l r r r r r}
\toprule
Task & AdamW epoch & MetaAdamW epoch & AdamW total & MetaAdamW total & Overhead \\
\midrule
ETTh1 & 93.39 & 99.04 & 747 & 693 & -7.20\% \\
WikiText-2 & 373.58 & 357.32 & 5604 & 4645 & -17.11\% \\
Multi30k & 895.25 & 957.67 & 18800 & 23942 & +27.35\% \\
CIFAR-10 & 2410.36 & 2635.86 & 28924 & 36902 & +27.58\% \\
IMDB & 283.35 & 356.40 & 1700 & 4633 & +172.53\% \\
\bottomrule
\end{tabular*}
\end{table*}

\section{Generative AI Usage Disclosure}
The authors used generative AI tools (DeepSeek) for language refinement and code adaptation. All scientific content, experimental design, and conclusions are the sole responsibility of the authors.

\section{Repository}
\href{https://github.com/qq150078158-lab/MetaAdamW}{https://github.com/qq150078158-lab/MetaAdamW}

\end{document}